\documentclass[journal, doublecolumn]{IEEEtran}

\usepackage{blindtext, graphicx}
\usepackage{amsmath}
\usepackage{amssymb}
\usepackage{subfigure}
\usepackage{cite}
\usepackage{amsxtra}
\usepackage{amsfonts}
\usepackage{graphicx}
\usepackage{multirow} 
\usepackage{listings}
\usepackage{amsmath}
\usepackage{caption}
\usepackage{blindtext}
\usepackage{rotating}
\usepackage{multirow}
\usepackage[table,xcdraw]{xcolor}
\usepackage{verbatim} 

\usepackage{colortbl}

\usepackage{hhline}
\usepackage[T1]{fontenc}
\usepackage[utf8]{inputenc}
\usepackage{tabularx,ragged2e,booktabs,caption}
\newcolumntype{C}[1]{>{\Centering}m{#1}}

\usepackage{chngcntr}
\usepackage{lipsum}
\usepackage[font=small,labelfont=bf]{caption}
\newenvironment{Table}
   {\par\bigskip\noindent\minipage{\columnwidth}\centering}
   {\endminipage\par\bigskip}

\ifCLASSINFOpdf
\else
\fi
\begin{document}

%
\title{IoT Security: Botnet detection in IoT using Machine learning}
%
%
%

\author{%
Satish Pokhrel, Robert Abbas, Bhulok Aryal
 \\ 
satish.pokhrel@students.mq.edu.au, robert.abbas@mq.edu.au ,bhulok.aryal@students.mq.edu.au

Macquarie University, Sydney, Australia
}

\maketitle

\begin{abstract}
  The acceptance of Internet of Things (IoT) applications and services has seen an enormous rise of interest in IoT. Organizations have begun to create various IoT based gadgets ranging from small personal devices such as a smart watch to a whole network of smart grid, smart mining, smart manufacturing, and autonomous driver-less vehicles. The overwhelming amount and ubiquitous presence have attracted potential hackers for cyber-attacks and data theft. Security is considered as one of the prominent challenges in IoT. The key scope of this research work is to propose an innovative model using machine learning algorithm to detect and mitigate botnet-based distributed denial of service (DDoS) attack in IoT network. Our proposed model tackles the security issue concerning the threats from bots. Different machine learning algorithms such as K- Nearest Neighbour (KNN), Naive Bayes model and Multi-layer Perception Artificial Neural Network (MLP ANN) were used to develop a model where data are trained by BoT-IoT dataset. The best algorithm was selected by a reference point based on accuracy percentage and area under the receiver operating characteristics curve (ROC AUC) score. Feature engineering and Synthetic minority oversampling technique (SMOTE) were combined  with machine learning algorithms (MLAs). Performance comparison of three algorithms used was done in class imbalance dataset and on the class balanced dataset.

\end{abstract}

\IEEEpeerreviewmaketitle

\section{Introduction}
Internet of things is an interconnection of billions of smart devices being able to communicate over the internet\cite{Chopra2019}. Over the past few years, the number of everyday machines embedded with sensors and can communicate over the internet are rising to a great extent. According to the article published in IoT Business News, Devices connected to the IoT world is increasing day by day which is expected to be around 24.1 billion by 2030\cite{Parker2005}. The internet of things makes the world smarter by merging the physical devices with the digital intelligence. International standardization sector (ITU-T) define the Internet of Things as an international structure consisting of interconnected devices based on information and communication technologies\cite{cite1}. There is large flow of data in between the interconnected devices and the security is the major concern in IoT.
\newline
\indent  As IoT connect the things to the internet and those things communicate with each other without any human involvement, IoT devices are susceptible to cyber-attacks of varying nature. To ensure the security of IOT network and devices proper security requirements should be identified at very beginning stages of design and deployment of IoT devices\cite{Dibaei2020}. As IoT concept is in an emerging phase, it still lacks robust security infrastructure/mechanism which create a risk on the valuable information. Modern security tactics must be adopted on IoT network to keep IoT entities, organizations, and individuals safe. The major security challenge in IoT is botnet-based DDoS attack where hackers infect the devices with script. 
\newline
\indent In this paper, a methodology for botnet detection is presented that comprises of data collection, data pre-processing, and SMOTE technique to balance the data-set class. Also, feature engineering was done upon analyzing machine learning algorithms for classification. We presented  the effect of imbalance data and its impact on machine learning.

\section{Related work}
\indent Botnet network is a sophisticated network of bots used by cyber criminals to launch malicious activities over the internet. Botnet based attack is one of the major challenges of IoT. Detection of attack in IoT network is notably distinct since it requires specific requirements for instance low latency, mobility, and distributed nature\cite{McDermott2018}. There are many researches which have been carried out to design effective botnet detection system, some of the previous work on botnet detection using different MLAs are discussed in this section.
\newline
\indent The paper presented by author Rudy Hartanto et.al\cite{Soe2019} proposed a botnet detection model based on artificial neuron network in which author implement data resampling technique that is SMOTE to resample real-time data into class balance data. Authors use BoT-IoT dataset with artificial neural network (ANN). Authors use ANN and SMOTE to model detection system. The proposed system is effective to detect DDoS attack with basic configuration with ANN.
\newline
\indent Author Muhammad Aamir et.al\cite{Aamir2019} proposed a machine learning model applying feature engineering to detect DDoS attacks. Authors combine feature engineering with different machine learning algorithms, k-nearest neighbors (KNN), Naive bayes (NB), support vector machine (SVM), Random forest (RF) and artificial neural network (ANN) to benchmark performance of different algorithms. Authors use feature selection methods of chi2 and information gain scores on supervised Machine learning algorithms (MLAs) for optimum tuning of features. The result obtained showed that the dimension of feature reduction is possible to improve the detection model performance. Authors validate the performance of proposed model with ROC AUC analyses and cross-validation technique. From evaluation on different MLAs author achieve best performance from KNN algorithms and analyses the accuracy scores of datasets with a smaller number of features. This paper demonstrate that feature reduction is possible with very less impact on accuracy, to reduce processing overhead in the system. Another similar research conducted by author\cite{Bahsi2018} on dimensionality reduction for Machine learning. Author stated that reducing the dimensionality of the required features overcome scalability and reduce computational overhead problems. In the paper author applied feature selection technique to select best features and use fewer features in decision tree algorithm. The paper present comprehensible results that verify that that fewer features helps to achieve higher accuracy rates.
\newline
\indent Author Vishwakarma et.al\cite{Dietz2018} proposed a honeypot-based botnet detection model that uses machine learning algorithms for botnet detection. Authors use IoT honeypot generated data to train machine learning model. Honeypot is used for tempting an attacker to study attack launching techniques by analyzing information about the malware that is used to launch botnet attack.  This information is used to train machine learning model. Authors use honeypot to allow attacker to gain access through open port with an intention to capture record of each activities between our device and attacker. This captured record gives the information about new malware families. Authors extract packet length, inter-packet intervals, and protocol features and implement them in different MLAs namely, random forest, K-nearest neighbors, SVM, decision tree and neural network and benchmark their performance. The advantage of this approach is the ability to detect new malware family used in botnet attack. 
\newline
\subsection{Research gap}
There are numerous research on botnet detection model, however, very few apply feature engineering so that complete evaluation would be possible to evade the problems with large datasets such as duplication and multicollinearity. Simply applying the traditional dataset without feature engineering may be problem solving but introduce over-fitting of the module.
\newline
\indent Most of the research are carried out with the traditional dataset which lack IoT-traces and they are not effective to detect modern botnet problem in IoT. Furthermore, most research carried out in the botnet detection model use real-time datasets which is highly imbalanced dataset. Researchers focus on finding high accuracy by using different MLAs on those imbalanced datasets. They overlook the effect of imbalance dataset in training module. The accuracy provided by that imbalanced dataset may be illusory.  Also, for evaluating the performance of MLAs researchers calculate accuracy after training algorithm. This gives the accuracy of training dataset and does not provide information about the skill of model on the unseen data. Accuracy percentage from such procedure may be illusory and there is possibility of over fitting or under-fitting of data. Furthermore, class imbalance dataset model is biased towards majority class, in such case accuracy might be impractical. 
\newline
\indent We combined the feature engineering and machine learning methods along with SMOTE resampling technique on real-time imbalanced dataset. We used latest released dataset i.e., BoT-IoT dataset,  created in IoT environment which includes DDoS attack traffic records. This dataset is also highly class imbalanced, so we used SMOTE technique to make it class balanced. Data training was done on all MLAs with both imbalanced datasets and class balanced datasets and the effect of imbalanced dataset in algorithms were analyzed and benchmarked the performance of algorithms to select best MLAs for our detection model. Also, to get clear information about the performance of our model, cross-validation technique was used that split the training data into number of folds (k) and for each time a subset of data is used as a test set and remaining (k-1) subset form a training set. This removed the problem of over-fitting or under-fitting of data and provided clear information about the performance of model for unforeseen data while training. Also, the model's performance was evaluated using ROC AUC curve and accuracy.

\section{Background Methodology}
In this section a brief descriptive overview of Botnet, security vulnerabilities in IoT, botnet malware, botnet life-cycle, different botnet detection methods and the concept of machine learning and machine learning algorithms used in this project is discussed.

\subsection{Botnet}
\indent Botnet is a network of numerous bots designed to perform malicious activities on the target network which are controlled using command and control protocol by the single unit called botmaster\cite{Almutairi2020}. Bots are the infected computers controlled remotely by the botmaster without any sign of being hacked and are used to perform malicious activities. Botnet size varies from small botnet consists of few hundred bots to the large botnets with 50,000 hosts. Hackers spread botnet malware and operate secretly without any noticeable indication of their presence and can remain effective and functioning for years. The botnet command and control architectures are shown in the Fig. 1.
\begin{figure}[h]
\captionsetup{justification=centering}
\includegraphics[height=4cm, width=8.5cm]{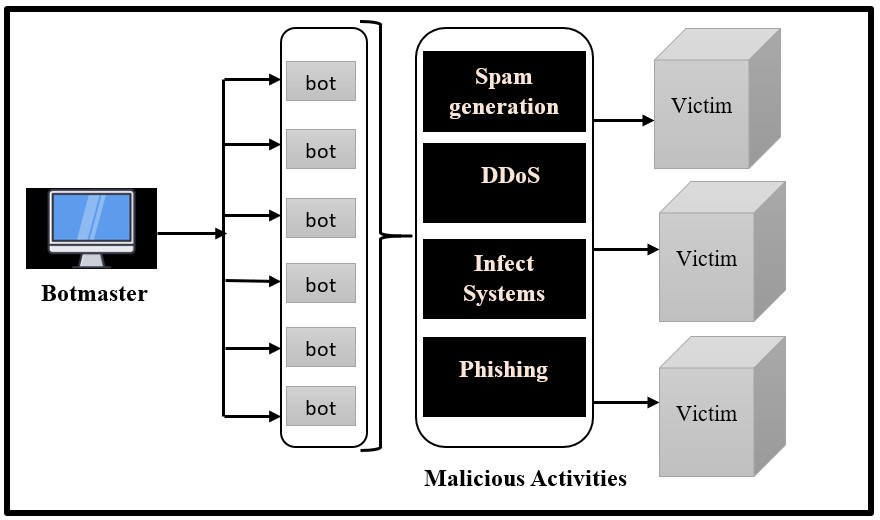}
\caption{Botnet Architecture}
\label{fig:fig3}
\end{figure}
The main element in botnet is the communication of botmaster with its associated bots. Communication with bots is essential to deliver commands to the bots to carry out malicious activities\cite{Vormayr2017}. Botmaster always stays hidden using low bandwidth and provide hidden services in the botnet network. Botmaster always communicate through command and control server with bots. The main goal of bots is to remain hidden until they are required to carry out assigned tasks.  The hidden nature of bots makes it harder to identify as they do not disrupt the normal operation on the host and remain silent until they receive command from botmaster to execute allocated activities.  The life cycle of botnet consists of several stages which involve spread and infection, secondary injection,  connection, malicious command and control, update, and maintenance\cite{Shetu2019}. 

\subsection{DDoS attack}
\indent DDoS attack is the most common cyber-attack in which attacker’s computers send large number of malicious traffic to the target server at the same time to overwhelms the target network\cite{Kansal2017}. DDoS attacks intend to significantly interrupt normal functioning of target server by flooding the target device with massive traffic such as fraudulent request to over saturate its capacity causing a disruption or denial of service to the legitimate traffic\cite{Dibaei2019}. DDoS attacks affect the server's system resources such as CPU, memory and can also cause the network bandwidth to saturate with large number of traffic, as a result, legitimate computers are going to be denied service because the server is pre occupied in dealing with DDoS attack. Hackers use botnet to launch DDoS attack. IoT devices get involved in DDoS attack after they gets infected by the malicious software that the attacker distributes over the internet.  Infected IoT devices acts as a bot and they are used by the attacker to launch DDoS attacker.
\begin{figure}[h]
\captionsetup{justification=centering}
\includegraphics[height=4cm, width=8.5cm]{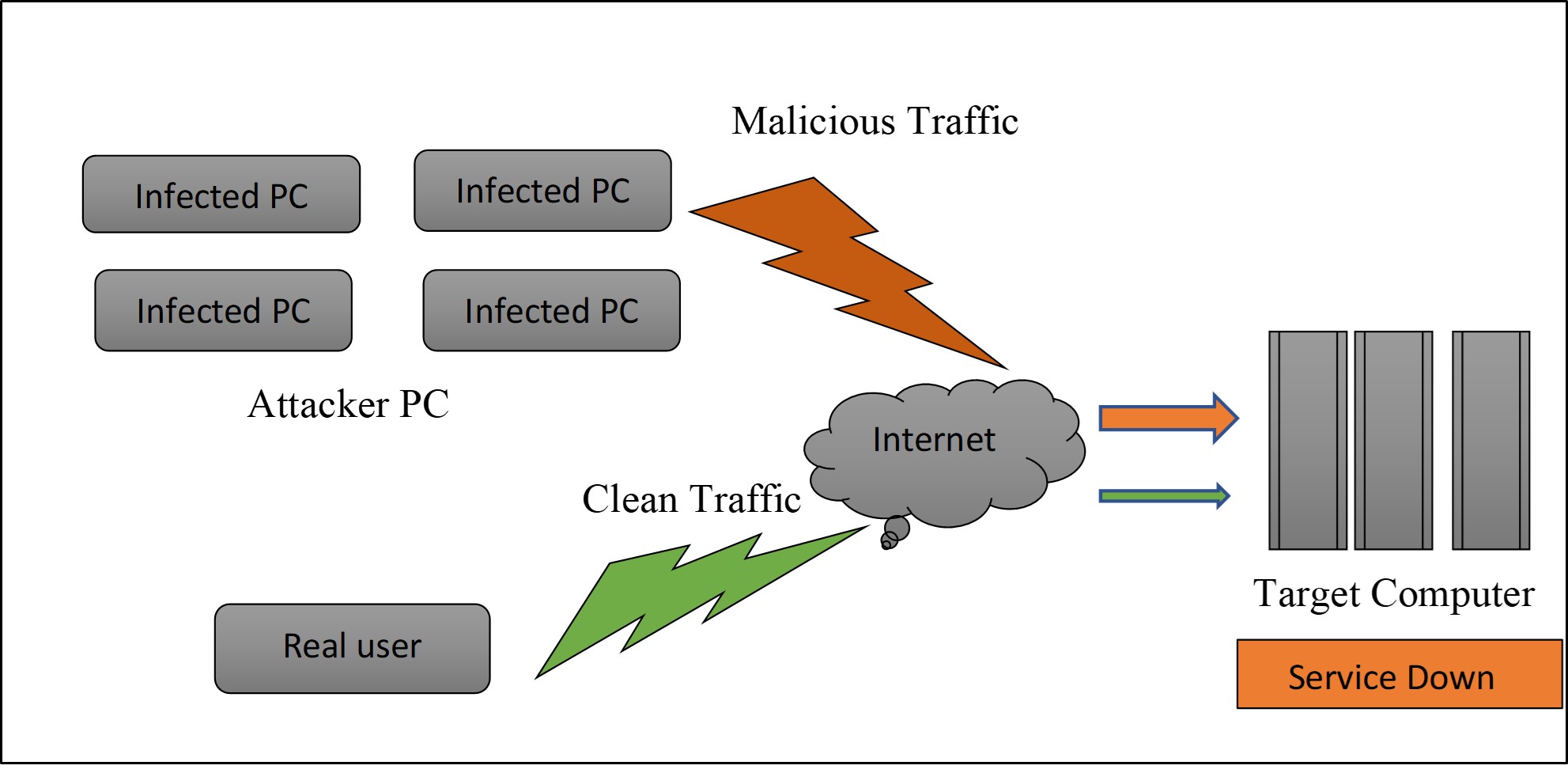}
\caption{DDoS Attack}
\label{fig:fig3}
\end{figure}

\subsection{Security vulnerabilities in IoT}
Smart devices are employed in various public and private sectors and its dynamically turning out to be basic objects of regular day to day living.This lead to high risk in data privacy. In such scenario, computerized security system utilizing the concept of Machine learning algorithms will be fated\cite{Meneghello2019}. Automated Security system using machine learning is critical to prevent threats like DDoS attack, Man-in-the-middle attack, botnet attack, eavesdropping and so on in an optimum way\cite{Section2020}. Also, most of the low-end IoT devices have weak security system, and henceforth are target or even used as a botnet for various security attack.

\section{Machine learning}
Current security systems greatly depend on mathematical models which often do not signify the systems accuracy and appropriate security in wireless environment requires hefty mathematical solutions which causes long computational time and create complexity\cite{Ali2020}. Machine learning algorithms will therefore undertake a crucial part in IoT security system, since it is effective in modelling systems that cannot be introduced by a mathematical equations.
Machine learning is the field of computer science that enable the machines to learn from previous examples and experience. Machine learning based anomaly detection system is the formation of groundbreaking new anomaly detection model to discover unusual traffic based on learning algorithm that might indicate an attempt of intrusions in the network\cite{Lam2020}. There are various machine learning algorithms (MLAs) to create a mathematical model formulated on sample data that add the ability on computers to decide without being explicitly programmed\cite{Hoang2018}. The MLAs are of three types of categories based on nature of supervision in training. They are supervised learning, unsupervised learning, semi-supervised learning, and reinforcement learning\cite{Ray2019}. 

\subsection{Supervised learning: }
\indent  In basic terms, supervised learning refers to learning approaches that enlist the assistance of supervisor. It consists of sample data labelled by defined outcome that simplifies the algorithm’s transition from input to output, as well as learning and prediction\cite{Schwitter2020}. Classification approaches such as KNN, SVM, Naive Bayes, Decision tree and Random forest, are all examples of supervised learning\cite{Shaheamlung2020}.

\subsection{Unsupervised learning: }
\indent Unsupervised learning refers to the method of analyzing with unlabeled data. It is also known as clustering. It is like self-guided learning process. The goal of unsupervised learning is to find the unusual data points \cite{Schwitter2020}.
\subsection{Semi-supervised learning:}
\indent Semi-supervised approach is a category of learning method in Machine learning that combines the limited amount of labeled data with larger set of unlabeled data \cite{Schwitter2020}. These learning falls somewhere between training data with labels and training data with no labels. These algorithms do better when dealing with large amounts of unlabelled data and fewer labeled data\cite{Shaheamlung2020}.

\subsection{Re-enforcement learning:}

\indent Reinforcement learning is the field of ML that based on agent, action, state, reward, and environment\cite{Shaheamlung2020}. It do not assume learning of any exact mathematical model, train an agent which consists of policy and learning algorithms through trial and error in an anonymous environment. 
\newline
\newline
\indent Each machine learning technique have their own application area. In this report, we implemented few supervised learning techniques in botnet detection.

\section{Supervised Machine learning algorithms}

\subsection{Gaussian Naïve bayes machine learning algorithm}
\indent Naïve bayes algorithm is one of the supervised machine learning algorithms that is founded on Bayes theorem. Bayes’ theorems apply the technique of maximum likelihood of case happening based on the previous learning\cite{Ray2019} . In simple language Bayesian probability can be expressed as:

\[ P(X/Y)=  (P(Y/X) P(X))/(P(Y)) \]
Where X and Y are occurrence and P(Y) is not equal to zero,
\newline
\indent P(X/Y): Likelihood of X case happening given that Y is true, 
\newline
\indent P(Y/X): Likelihood of Y case happening given that X is true,
\newline
\indent P(X) and P(Y) are likelihood of observing case X and Y, respectively.

\subsection {KNN}
K Nearest Neighbor algorithms works based on Euclidean distance calculation and object is categorized by majority of vote of its K neighbors with the entity of different classes\cite{Huang2018}. The value of K is positive and usually small. The accuracy of KNN algorithm depend on the number of neighbors chosen that is the value of K. Usually the value of K is chosen odd number for binary classification to evade the possibility of two classes labels acquiring the same count. If the value of K is chosen 1 then the entity is simply assigned to its single nearest class. Value of K chosen should be optimal, if the value of K is small, then it could be under-fitting as well as larger value can cause over-fitting of the model. The Euclidean distance between two points (X, Y) in Euclidean n-space is expressed as:

\[  d(X,Y) = \sqrt{\sum_{a=1}^n {(Ya-Xa)^2} }  \]

\subsection {MLP ANN}
Multilayer perception is an artificial neural network consisting of three layers of nodes which are input layer, hidden layer, and output layer\cite{Heidari2016}. Hidden and output nodes in MLP is a neuron that utilized nonlinear activation function. Nonlinear activation transfer functions are the mathematical equation which will be in different forms for example binary step function, Gaussian functions, identity function. MLP ANN is a supervised machine learning algorithm that uses back propagation technique. Each neuron in MLP is connected to its neighbors with variable weights. The weighted summation of the input is passed to the hidden neurons. Activation function in hidden neuron transformed the weighted sum and then passes to the output neuron. Again, in output neuron there undergoes another transformation by nonlinear activation transfer function in output layer and yields an outcome [38].

\section{Research approach}
In this section we describe the procedures followed during the botnet detection model creation including dataset used, data preprocessing, experimental scenario, results, and explanations. Different supervised MLAs were used on different combination of Botnet dataset and benchmarked the result to select a best algorithm for our model. At first, we analyzed the botnet behavior by examining the packet information. And the dataset was divided into two parts; one with normal traffic and other containing the botnet traffic and studied the behavior of botnet. This analysis helped to select features that contain more reliable information. Along with manual analysis we used feature selection and extraction technique to select appropriate features based on top \emph{F-score}. 
\begin{figure}[h]
\captionsetup{justification=centering}
\includegraphics[height=7cm, width=8.5cm]{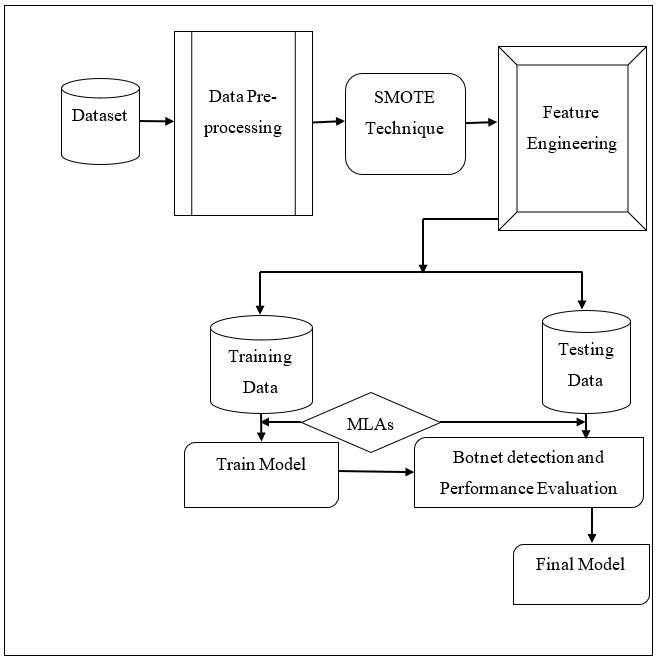}
\caption{Proposed model}
\label{fig:fig3}
\end{figure}

\subsection {Experimental dataset (BoT-IoT)}
Most of the available datasets lack recent attacks data and are not for IoT network. As our project is botnet detection in IoT environment, it requires dataset containing enough information about IoT traces. BoT-IoT dataset was created in the lab of UNSW Canberra cyber centre. This dataset collaborates the normal and botnet traffic with label. The researcher creates many virtual machines on internal network to simulate different malicious attacks with the intention to capture normal and malicious traffic. They capture more than 72 million records to create BoT-IoT dataset\cite{cite2}. The dataset includes traffic from different malicious attacks namely, DDoS, DoS, OS, Data exfiltration and Keylogging attacks with additional DDoS and DoS attacks set up on protocol used\cite{Koroniotis2019}. 
\begin{figure}[h]
\captionsetup{justification=centering}
\includegraphics[height=7cm, width=8.5cm]{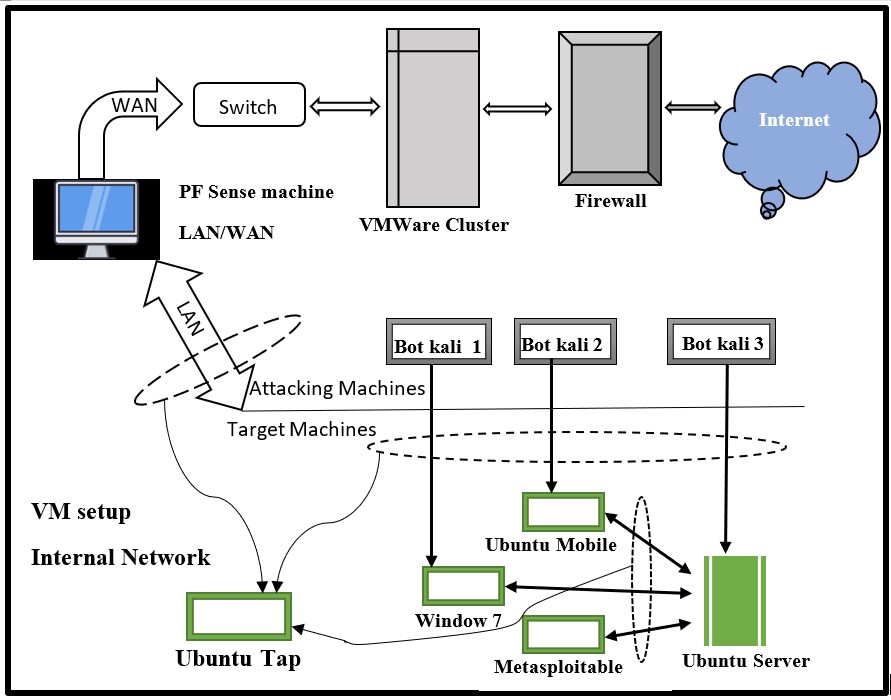}
\caption{Testbed environment of BoT-IoT Dataset}
\label{fig:fig3}
\end{figure}
BoT-IoT dataset is realistically designed in IoT network using different tools to create various botnet scenarios\cite{Koroniotis2019}. BoT-IoT dataset have realistic testbed and organized captured traffic based on attack types\cite{Soe2019}. BoT-IoT entire dataset contains 74 .csv files with altogether 72 million records with each file containing approximately 1 million records combining botnet and normal traffic. We selected one of the .csv files for my model preparation which contains 999,610 records with 994,828 botnet traffic and remaining normal traffic because the system we used took a lot of execution time in our device with limited system capability. The major contrast in this dataset is that it contains more than 99\% of botnet traffic while less than 1\% normal traffic. We created another dataset after processing real-time BoT-IoT dataset through SMOTE technique which provided class balance dataset with equal number of botnet traffic and normal traffic. 

\subsection{Data preprocessing}
Data preprocessing is the first and one of the most important stages in building machine learning model. Bad data can produce inaccurate result. To get correct output at very first stage, dataset is analyzed, and formatting is done. It involved data cleansing, normalization, and transformation to create reliable dataset. This enriched the data quality for training of machine learning module and supports precise decision-making. 
\subsubsection{Data Cleansing}
Data cleansing is the process to fix and remove incomplete information. We undergo through data cleaning process to identify missing values and delete those rows. We drop the rows containing null value in BoT-IoT dataset using dropna() function of pandas.
\subsubsection{Normalization}
Normalization is done to get common scale in the dataset. As some feature in our BoT-IoT dataset have data of variant range which make complex for the model to learn and will cause model learning problem taking it more time to decide to converge to result. Normalization helps the model to converge quickly, and this will increase the model performance. The normalization technique we used in our model is min-max feature scaling that transfer multiple scale feature to a fixed scale range of [0,1]. The mathematical expression of min-max feature scaling is:
\[ Y_norm=  (Y-Y_min)/(Y_max-Y_min ) \]
We get Ymin and Ymax by using .min() and .max() functions of pandas.

\subsubsection{Transformation}
Data transformation is the process of converting data from one format to another. In our BoT-IoT dataset there are many categorical features containing non-numeric data which needed to be converted into numeric format for the MLAs to process it as the MLAs we were using were in algebraic format. Data transformation was done to convert non-numeric data of categorical features into numeric format. The values in BoT-IoT dataset contain protocol types which is converted into numeric format by assigning each individual protocol type with numeric value.

\subsection{Feature Engineering}
Feature engineering technique is applied as a part of machine learning approach which helped in dimensionality reduction which thereby minimized the problem of over-fitting. This improved accuracy, and significantly reduced processing time. Also, it was  beneficial in selecting appropriate features that contain most important information about target variable and helped to improve the performance of model. Appropriate feature was selected based on the chi-square (chi2) value which is calculate by using the equation below. 
\[ Chi2= \sum(Ac-Ep)^2/Ep \]
Here, 
\newline
\indent Ac indicate actual value and Ep is expected value.
\newline 
\indent Chi2 value, also known as feature score (F-score) gives the discriminating power of feature for the prediction of target variable. 
\newline
\indent For this paper, we calculated f-score for each features. Only the features that have a f-score greater than mean value were used. We found out that there were eight features that have a f-score greater than mean value and those were selected for training and testing of model. 
\newline
\indent Feature engineering reduces complication in model and improve the performance of MLAs. Using all the features of dataset can cause over-fitting of model and can lead to excessive computing time which result in poor performance of model. 
\newline
\indent The detail of feature selection procedure is shown in Fig 5.
\begin{figure}[h]
\captionsetup{justification=centering}
\includegraphics[height=5cm, width=8.5cm]{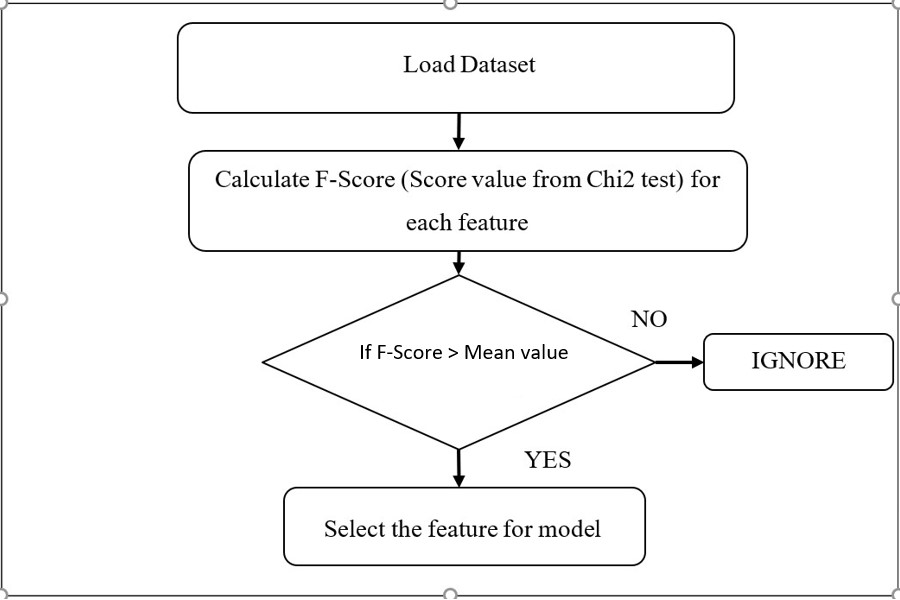}
\caption{Feature Engineering}
\label{fig:fig3}
\end{figure}

 \subsection{Synthetic Minority over-sampling technique (SMOTE)}
\indent SMOTE is one of the effective techniques to make class balance dataset. Imbalanced dataset may cause misclassifying problems which effect on the performance of machine learning algorithms. Synthetic Minority over-sampling technique over-sample the minority class. SMOTE technique chooses nearest neighbors, calculate the differences, and then multiply the difference value with random number between 0 and 1 to get more sample points\cite{Soe2019}. 
\begin{figure}[h]
\captionsetup{justification=centering}
\includegraphics[height=7cm, width=8.5cm]{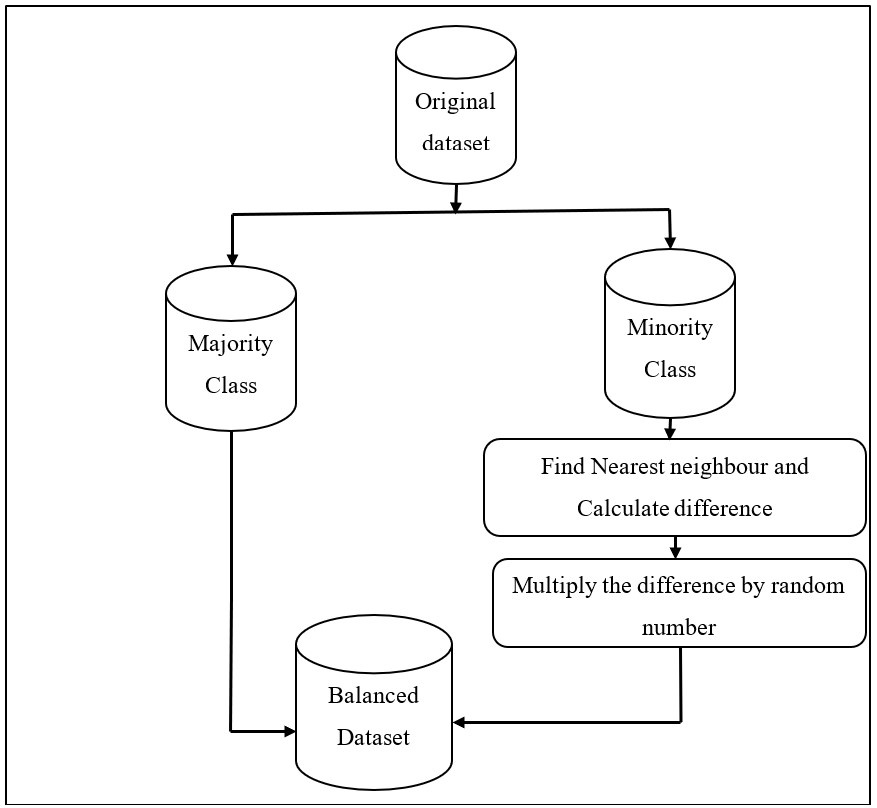}
\caption{SMOTE Technique}
\label{fig:fig3}
\end{figure}

\subsection{Experimental Scenario}
Hardware plays a vital role in the model performance. The system we used throughout the process of model creation and testing is constant. We used a laptop running windows 10, 64-bit operating system with 8GB RAM. The processor is 8th generation core i5 with 1.83GHz clock speed. The GPU is NVIDIA GeForce MX130. The storage is 488GB SSD.
\newline
\indent For training and testing, we used BoT-IoT dataset. The model is trained KNN, MLP ANN and Naive Bayes algorithms with the training datasets (D1 and D2) and the accuracy is the benchmarked to select the best algorithm for our detection system. Most of the real-word dataset are class imbalanced. The latest dataset for IoT with IoT traces is BoT-IoT dataset but it is highly imbalanced with more than 90 percent botnet traffic and just a few thousands normal traffic in each .csv file. To make our BoT-IoT dataset we used Synthetic Minority over-sampling technique also referred to as SMOTE. We implemented MLAs on datasets D1 and D2 and benchmarked performance of algorithm on unbalanced dataset(D1) and on class balanced dataset (D2). 
\newline
\indent For programming, we use Spyder platform running python 3.7. To process the dataset and implement machine learning, I use numerous python libraries.

\subsubsection{Python Libraries}
We mainly used Sklearn, imblearn, matplotlib and pandas.
\newline
•	Sklearn: 
Sklearn library is mainly used to create confusion matrix, to construct machine learning models, for splitting dataset, to perform data preprocessing and for feature engineering procedure.
\newline
•	Imblearn: 
Imblearn library support data sampling technology.  In this project we use Imblearn library to over sample minority class data to make dataset class balance.
\newline
•	Matplotlib:
Matplotlib library is used to visualize data in graphical format. This library support bar plot, scatter plot and many other plots that help in clear understanding of packet patterns.
\newline
•	Pandas:
Pandas library support data analysis. we use panda’s library to import dataset in .CSV file format and for data manipulating.

\subsection{Machine Learning Model Evaluation and Cross validation}
Performance evaluation of machine learning model is done by generating confusion matrix for each MLAs. We generate confusion matrix for each machine learning algorithms to gain insight into the type of error committed by machine learning model which help to understand the other metric such as accuracy that are derived from it. We derived accuracy, precision, recall and F1-Score from confusion matrix to evaluate the performance of model. The dataset used was  unbalanced and in case of unbalance dataset due to excessive variation in number of observations in different classes accuracy may lead to false result. In the dataset there are more samples of botnet traffic and very few normal traffic. Based on accuracy, precision, recall, F1-Score and ROC AUC, the performance of the model was evaluated.
\begin{Table}
\captionsetup{justification=centering}
\captionof{table}{Table layout of confusion matrix}
\includegraphics[width=8cm]{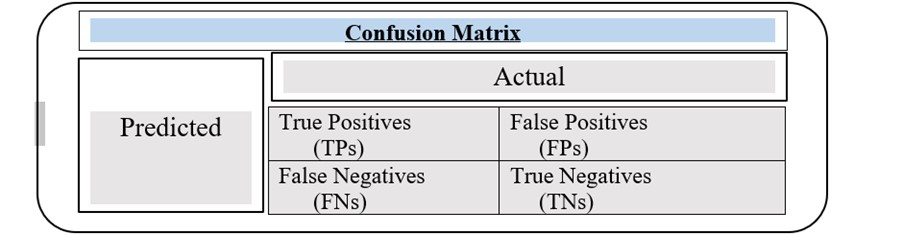}
\end{Table}
\subsubsection{Accuracy:}
 \indent Accuracy is the ratio of number of correctly predicted class to the total predictions. It’s presented in percentage. Accuracy is analyzed when True Positives (TPs) and True Negative (TNs) are crucial.
\[Accuracy=((TPs+TNs)/(TPs+TNs+FPs+FNs)*100)\% \]
\subsubsection{Precision:}
 \indent Precision is the ratio of number of correctly predicted positives to the total predicted positives. Precision is analyzed to minimized false positives.
\[ Precision=( TPs/(TPs+FPs )*100)\% \]
\subsubsection{Recall:}
\indent Recall is the ratio of number of correctly predicted positives to all positive examples. Recall is analyzed to minimized false negatives.
\[ Recall= (TPs/(TPs+FNs)*100)\% \]
\subsubsection{F1-Score:}
\indent F-Score metric combine precision and recall giving a single score value to balance both the concerns of precision and Recall. F1-Score is analyzed when false positives and false negatives are important.
\[ F1 Score=((2*precision*Recall)/(Precision+Recall)*100)\% \]

\subsubsection{ROC AUC:}
 \indent ROC stands for receiver operating characteristics and AUC is Area under ROC curve. ROC AUC is popular among class imbalanced dataset. My BoT-IoT real dataset is highly imbalance with 994,828 Botnet traffic and 4728 Normal traffic in 999,556 total data. This is 99.527 percentage botnet traffic and just 0.473 percentage Normal traffic. Classifier always predicting each traffic as botnet traffic will still have more than 90 percentage accuracy. So, for effective evaluation of model, analysis of ROC AUC curve is prefered. ROC is a graph with X-axis false positive rate (FPR) and Y-axis true positive rate (TPR). 
\[ FPR=FP/((FP+TN)) \]
\[ TPR=TP/((TP+FN)) \]
In this project we were using millions of data to train the model. Splitting entire data to train and test is not enough to give complete information about the performance of our system. Cross-validation technique splits the training dataset into multiple subsets which is called folds(k). The number of subsets depend on the value of k chosen. The value of k should be optimum. Higher value of k means more variance and less bias but may cause computational overhead and require more time whereas low value means low variance and high bias. Generally, value of k is selected 5 or 10. In our project we choose 5. cross-validation is simply a re-sampling procedure. In cross-validation technique each time a subset of data is used as a test set and remaining(k-1) subset form a training set. This technique gives much more information about the MLAs performance and mitigate the problem of over-fitting. It gives assurance about the stability of model performance and most important its low on bias and variance. Commonly, accuracy calculation is done after training which estimate the skill of Machine learning model on training data. There is possibility of overfitting or underfitting of data. Traditional performance evaluation technique does not give any knowledge about the model performance on unseen data set. Cross validation technique provides skill of model on real data.

\section{Experiment and Discussion}
\indent To benchmark the performance of MLAs, we implemented classifier algorithms on two set of BoT-IoT dataset, D1 and D2. D1 is real time BoT-IoT dataset where D2 is class balanced dataset created by using SMOTE technology in real dataset.  
The very first step in experiment is analysis of dataset. Entire dataset is divided into two parts i.e., normal traffic and botnet traffic to observe the count of normal and botnet traffic in the dataset. It can be more clarified from Fig 7 and Fig 8..
\begin{figure}[h]
\captionsetup{justification=centering}
\includegraphics[height=3cm, width=7.5cm]{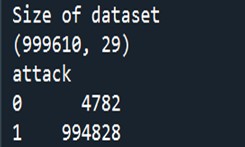}
\caption{Size of class in dataset}
\label{fig:fig3}
\end{figure}
\newline
\begin{figure}[h]
\captionsetup{justification=centering}
\includegraphics[height=5cm, width=8.5cm]{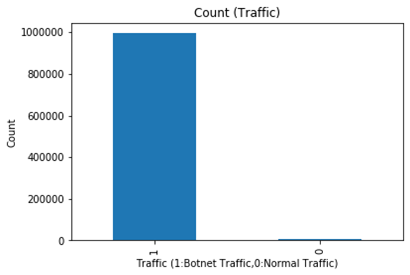}
\caption{Bar graph showing class imbalance in data-set}
\label{fig:fig3}
\end{figure}
\newline
\indent The Fig. 7 show that the dataset consists of 29 columns (features) and 999,610 rows. Among 999,610 data, 4782 belongs to normal traffic and remaining 994,828 belongs to botnet traffic. It shows that the dataset is highly imbalance.
Then data preprocessing was done to get reliable data. In order to get this I drop the rows containing null values using dropna() function of pandas.
As this dataset contains many non-numeric values. We assign numeric values to protocol names and state values. For proto feature, we assign numbers starting from 1 and for state feature we assign each value with a number starting from 10.

\subsection{Transformation}
The below picture shows the first 5 rows of the dataset before transformation.
\begin{figure}[h]
\captionsetup{justification=centering}
\includegraphics[height=3.5cm, width=8.5cm]{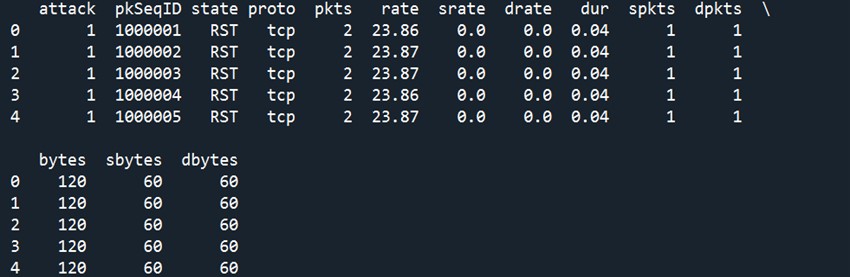}
\caption{First 5 rows of dataset before transformation}
\label{fig:fig3}
\end{figure}
\newline
\indent The below picture shows the first 5 rows of the dataset after transformation.
\begin{figure}[h]
\captionsetup{justification=centering}
\includegraphics[height=3.5cm, width=8.5cm]{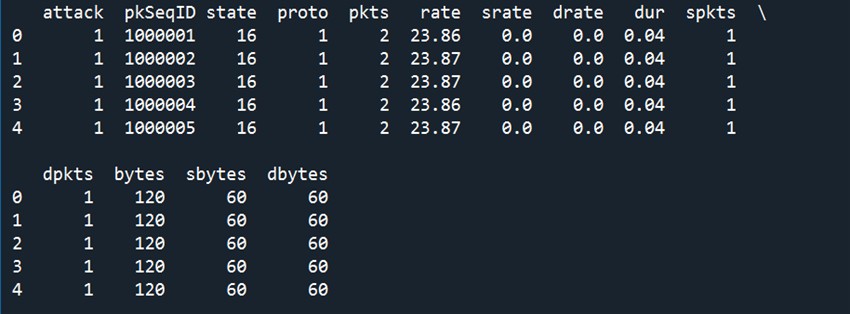}
\caption{First 5 rows of dataset after transformation}
\label{fig:fig3}
\end{figure}
\subsection{Analyzing the behavior of Normal and Botnet Traffic} 
\indent We analyze the statistics of botnet traffic and normal traffic to get information about the behavior of botnet.
\begin{figure}[h]
\captionsetup{justification=centering}
\includegraphics[height=4cm, width=8.5cm]{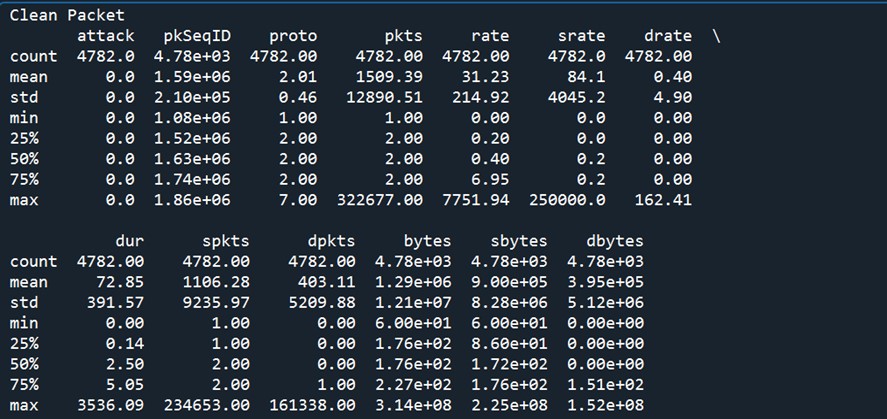}
\caption{Statistics of Normal Traffic.}
\label{fig:fig3}
\end{figure}
\newline
\begin{figure}[h]
\captionsetup{justification=centering}
\includegraphics[height=4cm, width=8.5cm]{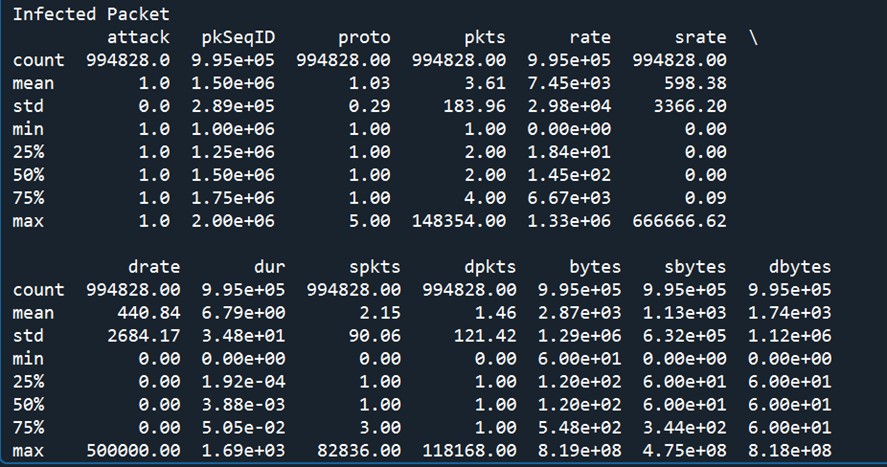}
\caption{Statistics of Botnet Traffic}
\label{fig:fig3}
\end{figure}
\newline
\indent Mean values of few important features of Normal and botnet traffic are presented below in the tabular form:
\begin{figure}[h]
\captionsetup{justification=centering}
\includegraphics[height=4cm, width=8.5cm]{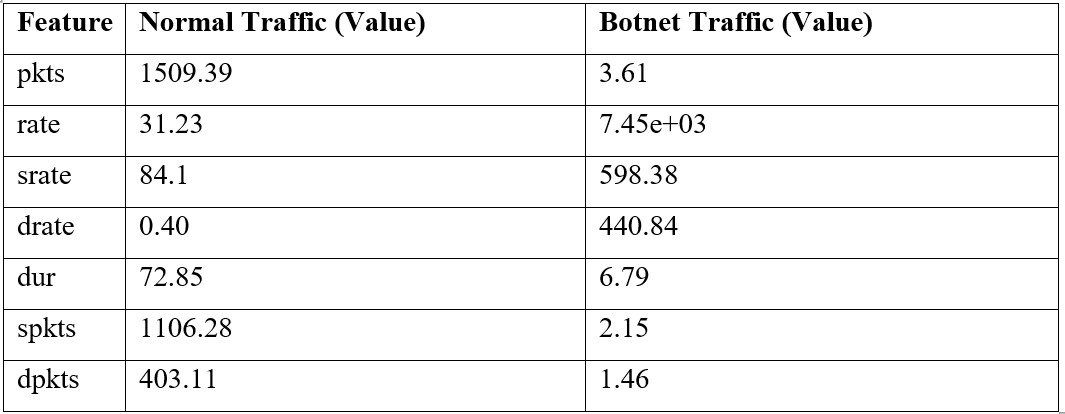}
\caption{Mean values of few important features of Normal and Botnet traffic}
\label{fig:fig3}
\end{figure}
\newline
\indent Major distinguishing behavior observed from statistics of Normal and botnet traffic:
\subsubsection{Total count of packets in transaction (pkts):}
\indent Mean value of \emph{pkts} of Normal traffic is 1509.39 while for botnet traffic its 3.61. This shows that the number of packets in transaction of botnet traffic is very less. This indicate that botnet traffic uses small size and a smaller number of packets in the transaction, which are not encrypted by most secure protocol. Normal traffic contains large payload content thus have large packet size. Average value of MTU in internet is 1400bytes.
\subsubsection{Total packet per second in transaction (rate):} 
\indent Mean value of rate of Normal traffic observed is 31.23 while of botnet traffic is 7.45e+03. This indicate rate of flow of packet of botnet traffic is much higher than normal traffic. 
\subsubsection{Source to destination packet per second (srate):}
\indent Mean value of \emph{srate} of Normal traffic and botnet traffic observed are 84.1 and 598.38. This indicate source to destination packet flow per second of botnet traffic is higher than of normal traffic.
\subsubsection{Destination to source packet per second (drate):}
\indent Mean value of \emph{drate} of Normal traffic and botnet traffic observed are 0.40 and 440.84. This indicate destination to source packet flow per second of normal traffic is very low while of botnet traffic is high. 
If we analyze \emph{srate} and \emph{drate}, source to destination packet rate and destination to source packet rate of botnet traffic, we see that there is not much difference in the packet flow rate from source to destination and vice-versa. 
\subsubsection{Record total duration (dur):}
\indent Mean value of ‘\emph{dur}’ of normal and botnet traffic observed are 72.85 and 6.79, respectively. This indicate that record total duration of packet exchange of normal traffic is higher than botnet traffic. From this mean value it is clear that attacker send large number of packets in short duration time to attack on the target.
\subsubsection{Source to destination packet count (spkts):} 
\indent Mean value of \emph{spkts} of normal and botnet traffic observed are 1106.28 and 2.15, respectively. This indicate that number of sources to destination packet of normal traffic is higher than of botnet traffic.
\subsubsection{Destination to source packet count (dpkts):}
\indent Mean value of \emph{dpkts} of Normal and botnet traffic are 403.11 and 1.46. This indicate that \emph{dpkts} of normal traffic is larger than of botnet traffic. 

\subsection {Oversampling minority class}
The below pictures illustrate the implementation of SMOTE technique used to over-sample minority class.
\begin{figure}[h]
\captionsetup{justification=centering}
\includegraphics[height=3.5cm, width=8.5cm]{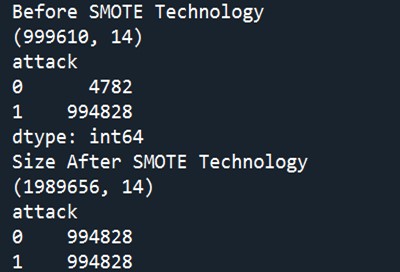}
\caption{Size of dataset before and after SMOTE technique.}
\label{fig:fig3}
\end{figure}
\begin{figure}[h]
\captionsetup{justification=centering}
\includegraphics[height=4cm, width=8.5cm]{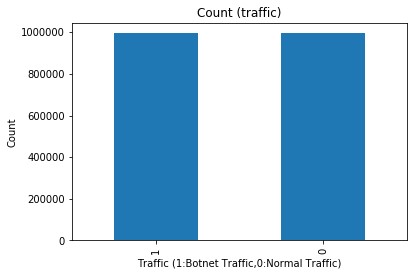}
\caption{Bar graph showing class balanced dataset after SMOTE technique}
\label{fig:fig3}
\end{figure}
\newline
\indent Real time BoT-IoT dataset contains 999,610 data, out of which 4782 belong to Normal traffic and 994,828 belong to Botnet traffic. After processing data through SMOTE technique, we get 1,989,656 data containing equal number of botnet and Normal traffic, that is 994,828. This makes the dataset class balanced.  
\subsection{Feature Score}
After analyzing the dataset, feature engineering was used to select the correct features based on the calculated feature score value for this model which provide the feature importance in distinguishing the classes. Top 8 features in accordance with the feature engineering in accordance with feature scores are presented below.
\begin{figure}[h]
\captionsetup{justification=centering}
\includegraphics[height=6cm, width=7cm]{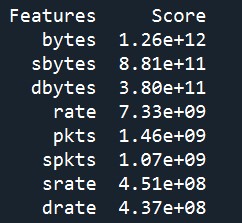}
\caption{Top 8 respective feature score}
\label{fig:fig3}
\end{figure}
\indent \emph{bytes, sbytes, dbytes, rate, pkts, spkts, srate} and \emph{drate} are the top 8 features based on chi2 value that were used to train the model.
\subsection{Train-Test Split}
\indent The dataset was splitted into train and test to evaluate the performance of model. 80 percent of data were used for training and 20 percent for test.
\subsection{Comparison of performance of Machine learning algorithms}
\indent  In this Botnet detection model, we combined feature engineering, SMOTE technology, and machine learning algorithms. In addition to that, we analyzed the performance of algorithms using cross-validation technique which divides the training set into number of subsets which gives real performance of model. The dataset we used in this model has class imbalance containing less than 1\% Normal traffic while there are more than a 99\% botnet traffic. The problem with unbalanced dataset is MLAs have poor predictive performance on minority class. If the training data is highly unbalanced, the algorithms may predict the majority class without examining comprehensive features. It may still have higher accuracy which is illusive. Imbalanced data problems can be mitigated by simply oversampling the minority class by duplicating data of minority class. However, this does not add additional information to the system. Synthetic minority Oversampling technique (SMOTE) is one of the simpler and effective approach to oversample imbalanced datasets. For the evaluation of this model, we used the cross-validation method. We evaluated true positive rate, false positive rate, precision, and recall.

\subsubsection{Results with Gaussian Naïve bayes model}

\indent As seen in Fig 17, for real BoT-IoT dataset, with Gaussian Naïve bayes algorithm, we observed nearly 100\% accuracy but 60.9\% ROC-AUC, and very low value in recall and f1-score. This evidently showed that accuracy is generally not helpful in imbalanced data. As we have more than 99 percent botnet traffic and less than 1 percent normal traffic in this dataset, this classifier may possibly classify all samples as 1. Thus, we get high accuracy but low ROC AUC (around 60 percent).
\newline
\indent Fig 18 shows ROC AUC graph of Gaussian model obtained from real-time dataset and class balanced dataset.
\begin{figure}[h]
\captionsetup{justification=centering}
\includegraphics[height=4.5cm, width=9.5cm]{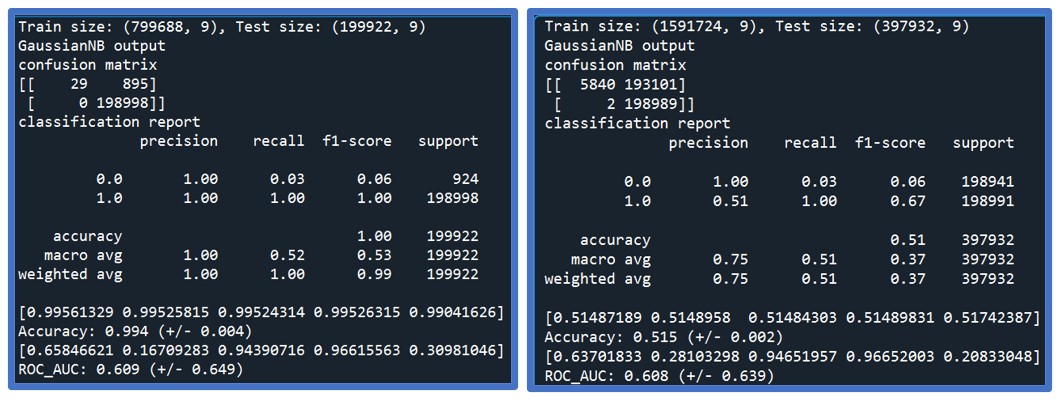}
\caption{Gaussian Naïve bayes model performance}
\label{fig:fig3}
\end{figure}
\begin{figure}[h]
\captionsetup{justification=centering}
\includegraphics[height=4.5cm, width=9.5cm]{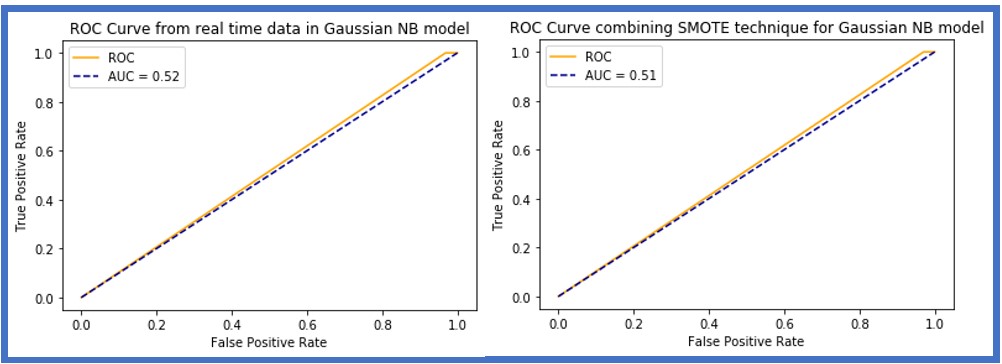}
\caption{ROC AUC graph from Gaussian NB model}
\label{fig:fig3}
\end{figure}
\newline
\indent From the ROC-AUC curve it is clearly seen that AUC score from Gaussian is 0.51. This denotes the gaussian algorithm is not effective to distinguish botnet and normal traffic.
\subsubsection{Results with KNN model}
\begin{figure}[h]
\captionsetup{justification=centering}
\includegraphics[height=4.5cm, width=8.5cm]{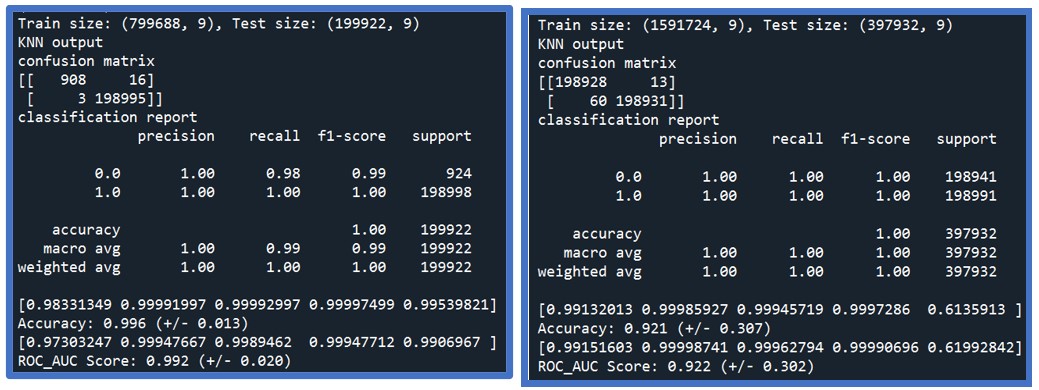}
\caption{KNN model performance}
\label{fig:fig3}
\end{figure}
\indent Fig. 19 shows KNN model performance on real-time class imbalance data and Class balance dataset is good. The accuracy and ROC AUC we get on BoT-IoT unbalanced data is 99.6\% and 99.2\% respectively and from class balanced data is 92.1\% and 92.2\% respectively. For the BoT-IoT dataset, the accuracy result from KNN is good. This indicates that the KNN classifier is effective algorithm in botnet detection system.
\begin{figure}[h]
\captionsetup{justification=centering}
\includegraphics[height=4.5cm, width=8.5cm]{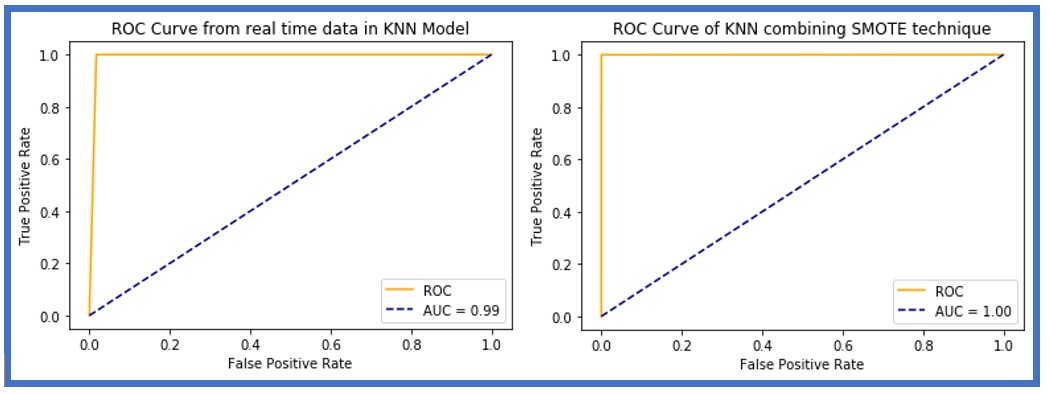}
\caption{ROC AUC graph from KNN model}
\label{fig:fig3}
\end{figure}
\newline
\indent ROC AUC curve of KNN model for botnet imbalanced dataset and class balanced dataset is almost 1 as seen in Fig 20. This indicates that KNN model effectively distinguishes botnet and normal traffic. KNN is better algorithm to use in botnet detection system.
\subsubsection{MLP ANN model}

\indent As presented Fig. 21, it is clearly seen that MLP ANN accuracy is 87.4 percent but low value in precision, recall, f1-score, and ROC AUC score. This showed that the accuracy value we got from the class imbalanced dataset is illusory. After integrating SMOTE technology, we got better value in precision, recall, f1-score, and ROC AUC. This validate that accuracy percentage is not enough to validate the model performance. Accuracy that we got from MLP ANN model combining SMOTE technique in BoT-IoT dataset is 85.8. This validates that MLP ANN algorithm is better to classify botnet and normal traffic. 
\begin{figure}[h]
\captionsetup{justification=centering}
\includegraphics[height=4.5cm, width=8.5cm]{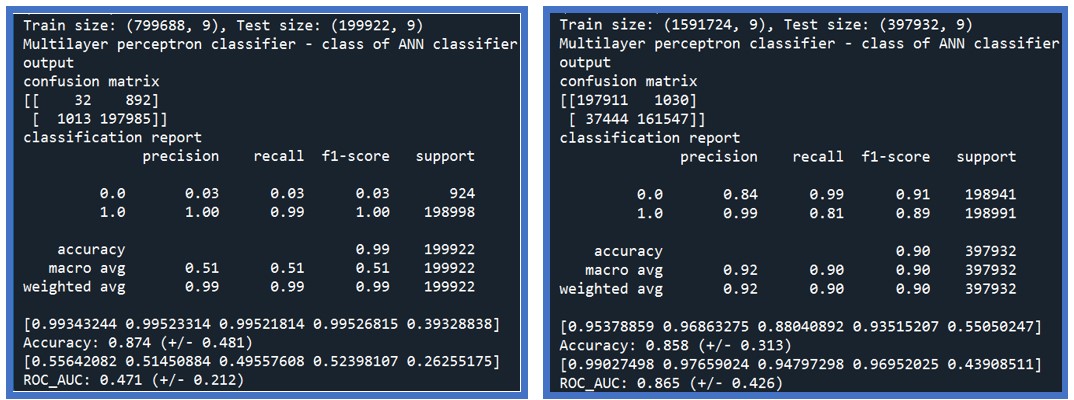}
\caption{MLP ANN model performance}
\label{fig:fig3}
\end{figure}
\begin{figure}[h]
\captionsetup{justification=centering}
\includegraphics[height=4.5cm, width=8.5cm]{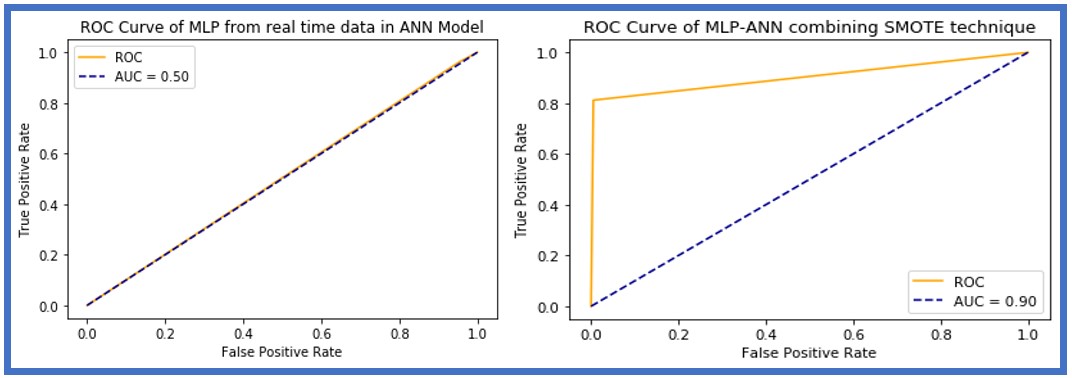}
\caption{ROC AUC graph showing MLP ANN model performance}
\label{fig:fig3}
\end{figure}
\newline
AUC score from MLP ANN before implementing SMOTE technique was 0.50 and after combining SMOTE technique on BoT-IoT dataset was 0.90 which can be seen in Fig. 22. This shows that MLP ANN is not effective enough on highly class imbalanced dataset. MLP ANN is good on class balanced dataset. It is also effective on botnet detection system.

\subsection{Observations}
While going through the process of botnet detection using gaussian ML algorithm we observed unexpected, and even surprising, results. We got accuracy of 99\% but the recall and ROC AUC scores were very low. After getting this undesired result, we analyzed the model performance on imbalanced and class-balanced data by implementing SMOTE technique on class imbalanced dataset and used feature selection method to reduce the dimension of feature to reduce the computation overhead of machine learning algorithm. Instead of using all the features, we used features with high feature score that improved the algorithm accuracy as well as reduced computational overhead. 
Table below shows the results from different MLAs on two set of datasets, D1 and D2. 
D1: Real time BoT-IoT dataset 
D2: Dataset obtained after processing BoT-IoT dataset through SMOTE technique.
\begin{Table}
\captionsetup{justification=centering}
\captionof{table}{Comparison of MLAs performance}
\includegraphics[width=8cm]{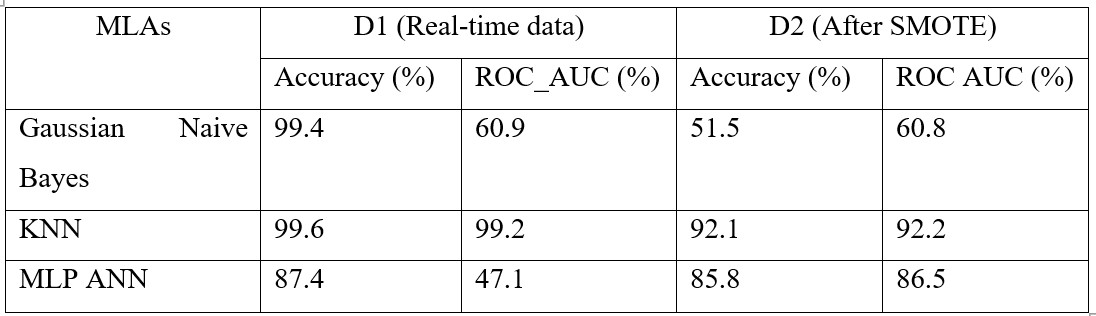}
\end{Table}
\begin{figure}[h]
\captionsetup{justification=centering}
\includegraphics[height=4.5cm, width=8.5cm]{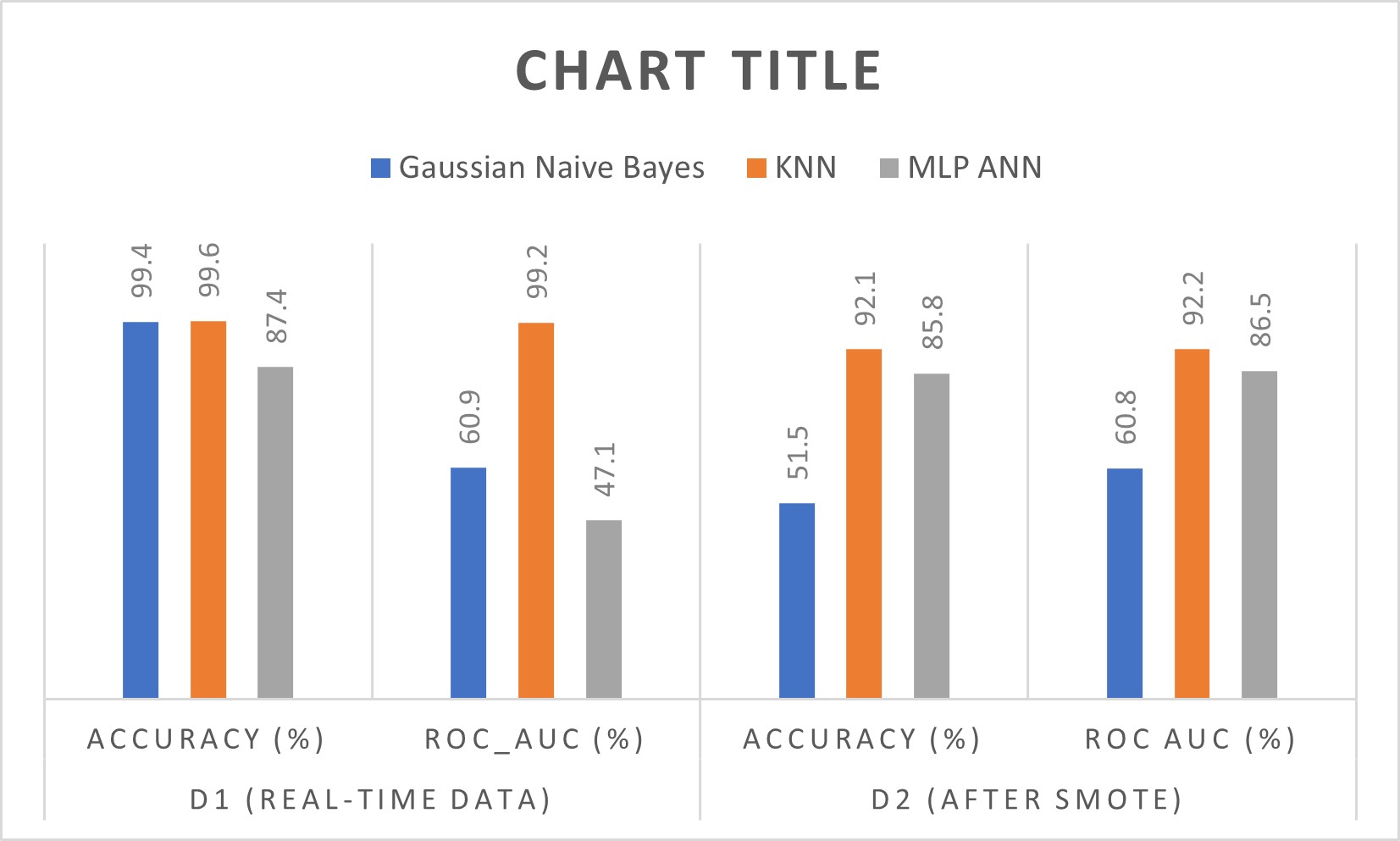}
\caption{Graphical presentation of machine learning performance comparison}
\label{fig:fig3}
\end{figure}

\indent From the above comparison, we get good and stable accuracy from KNN model. We got 92.1\% accuracy and 92.2\% ROC AUC from KNN algorithm. Also, KNN algorithm works good on highly imbalanced real-time data. KNN is effective to use in botnet detection system. Among different machine algorithms we get higher accuracy from KNN algorithm. From these overall comparisons on different evaluation metrics of machine learning algorithms, KNN algorithm was found to be the best for BoT-IoT dataset.

\section{Conclusion}
\indent  In this paper, we propose K-nearest neighbors’ algorithm as an effective botnet detection model and its performance is evaluated against available methods and algorithms to detect and mitigate botnet-based DDoS attack in IoT network. Our comparison of botnet detection models on real time imbalanced dataset and balanced dataset considerably help to enrich our research. It revealed us how and why the real time imbalanced datasets were not optimum, how it affects the metrics such as precision, recall, accuracy, f1-score, and ROC AUC and how the dataset should be improved. The usage of imbalanced dataset, despite showed us the good accuracy, recall and f1-score were low. This shows that accuracy we got from imbalanced dataset maybe illusory. After combining SMOTE technology, we got more stable accuracy and ROC AUC with similar range of values on precision, recall and f1-score. This proves that with implementation of SMOTE technology we can get more reliable performance of the model. Based on the findings, KNN algorithm was found to be the most reliable in botnet detection.

\section{Future work}

\indent  Future work of this paper would concentrate on simulation of the proposed model for real-time comparison for effectiveness. This model can be implemented with Software Defined Network (SDN). Addition of botnet detection and mitigation measure that detect the botnet and block the host that sending botnet packet over the network can be used on the controller of SDN. This will help to monitor traffic flow in all the connected host and thereby effectively mitigate the botnet attacks in server.

\bibliographystyle{ieeetr}
\bibliography{biblography}

\end{document}